%% file: main.tex
\documentclass[10pt]{article}

\input{preamble.tex}

\usepackage{lmodern}

\usepackage{graphicx}
\usepackage{times}

\usepackage{graphicx}
\usepackage{amsmath,amssymb} 
\usepackage{color}

\usepackage{tikz}
\usetikzlibrary{backgrounds,calc,trees,hobby}
\usetikzlibrary{shapes,decorations}
\usetikzlibrary{arrows}
\usepackage{mathpazo}

\usepackage[pagebackref=true,breaklinks=true,colorlinks,bookmarks=false]{hyperref}

\usepackage[caption=false]{subfig}

\usepackage{xspace}
\newcommand{\latinphrase}[1]{\textit{#1}}  
\newcommand{\etal}{\latinphrase{et~al.}\xspace}

\newcommand{\figref}[1]{\mbox{Fig.~\ref{#1}}}

\newcounter{row}
\newcounter{col}

\tikzstyle{horizontalGradient} = [draw, fill=blue!50,shape=diamond,minimum size=4.25mm]
\tikzstyle{verticalGradient}   = [draw, fill=red!50,shape=rectangle,minimum size=3mm]
\tikzstyle{PixelNode}    	   = [draw, fill=yellow!90,shape=circle,inner sep=0.5pt,minimum size=3mm]

\tikzstyle{horizontalGradientLight} = [draw, fill=blue!10,shape=diamond,inner sep=1pt,minimum size=4.25mm]
\tikzstyle{verticalGradientLight}   = [draw, fill=red!10,shape=rectangle,inner sep=1pt,minimum size=3mm]
\tikzstyle{PixelNodeLight}    		= [draw, fill=yellow!30,shape=circle,inner sep=0.5pt,minimum size=3mm]

\tikzstyle{HighLightBlue} = [draw, fill=blue!100,shape=circle,inner sep=0.5pt,minimum size=3mm]
\tikzstyle{HighLightRed}  = [draw, fill=red!100,shape=circle,inner sep=0.5pt,minimum size=3mm]

\definecolor{myPink}{RGB}{255,182,193}
\definecolor{myRed}{RGB}{220,20,60}
\definecolor{myOrange}{RGB}{255,140,0}
\definecolor{myLightGreen}{RGB}{192,255,62}
\definecolor{myGreen}{RGB}{48,128,20}
\definecolor{myGray}{RGB}{205,205,205}
\definecolor{myViolet}{RGB}{205,50,120}

\tikzstyle{TreePink} = [draw,fill=myPink,shape=circle,inner sep=0.5pt,minimum size=2.5mm]
\tikzstyle{TreeRed} = [draw,fill=myRed,shape=circle,inner sep=0.5pt,minimum size=2.5mm]
\tikzstyle{TreeOrange} = [draw,fill=myOrange,shape=circle,inner sep=0.5pt,minimum size=2.5mm]
\tikzstyle{TreeLightGreen} = [draw,fill=myLightGreen,shape=circle,inner sep=0.5pt,minimum size=2.5mm]
\tikzstyle{TreeGreen} = [draw,fill=myGreen,shape=circle,inner sep=0.5pt,minimum size=2.5mm]
\tikzstyle{TreeGray} = [draw,fill=myGray,shape=circle,inner sep=0.5pt,minimum size=2.5mm]
\tikzstyle{TreeBox} = [draw,rounded corners,shape=rectangle,inner sep=0.5pt,minimum size=3.5mm]

\tikzstyle{horizontalGradientSmall} = [draw, fill=blue!50,shape=diamond,minimum size=4.25mm]
\tikzstyle{verticalGradientSmall}   = [draw, fill=red!50,shape=rectangle,minimum size=3mm]

\usepackage{pgfplots}
\usepackage{pgf,tikz}
\usepackage{mathrsfs}
\usetikzlibrary{calc}
\usetikzlibrary{arrows}
\pgfplotsset{compat=1.14}

\definecolor{colorMSHR}{RGB}{255,0,0}
\definecolor{colorMSER}{RGB}{255,255,0}
\definecolor{colorBox}{RGB}{0,255,255}

\def\addlegendimage{\csname pgfplots@addlegendimage\endcsname}
\def\addlegendentry{\csname pgfplots@addlegendentry\endcsname}

\begin{document}
\noindent

\bibliographystyle{plain}

\title{Efficiently Tracking Homogeneous Regions in Multichannel Images}

\authorname{Tobias B\"ottger, Christina Eisenhofer}
\authoraddr{MVTec Software GmbH, Germany, {\tt \{boettger,eisenhofer\}@mvtec.com}}


\maketitle


\abstract
We present a method for tracking Maximally Stable Homogeneous Regions (MSHR) in images with an arbitrary number of channels. MSHR  are conceptionally very similar to Maximally Stable Extremal Regions (MSER) and Maximally Stable Color Regions (MSCR), but can also be applied  to hyperspectral and color images while remaining extremely efficient. The presented approach makes use of the edge-based component-tree which can be calculated in linear time. In the tracking step, the MSHR are localized by matching them to the nodes in the component-tree. We use rotationally invariant region and gray-value features that can be calculated through first and second order moments at low computational complexity. Furthermore, we use a weighted feature vector to improve the data association in the tracking step. The algorithm is evaluated on a collection of different tracking scenes from the literature. Furthermore, we present two different applications: 2D object tracking and the 3D segmentation of organs.

\section{Introduction}
Extracting Maximally Stable Extremal Regions (MSER) \cite{matas2004robust} is an essential step in many image processing applications. The possible applications range from stereo feature point extraction \cite{matas2004robust} to optical character recognition (OCR) \cite{neumann2012real}. The linear time MSER algorithm \cite{nister2008linear} has also been extended to track extremal regions in gray-scale images by Donoser and Bishop \cite{donoser2006efficient}. The approach makes extensive use of the gray-scale component-tree, which is used to efficiently localize the MSER regions. Furthermore, as opposed to many of the existing tracking schemes, their approach does not represent objects by a bounding box, but rather by a dense segmentation. This allows the object to undergo non-linear deformations and still be tracked efficiently. 

The success of MSER for gray-scale images, and the increasing demand for image processing techniques devoted to color, has motivated the extension of MSER to multichannel images. A well known extension are the so-called Maximally Stable Color Regions (MSCR) \cite{forssen2007maximally}. Although conceptually very similar to MSER, no component-tree is constructed in the segmentation process. Hence, the efficient extension of MSER to multiple channels is not possible using the MSCR framework.

\begin{figure}
\begin{center}
\subfloat[][]{\fbox{\includegraphics[width=0.09\textwidth]{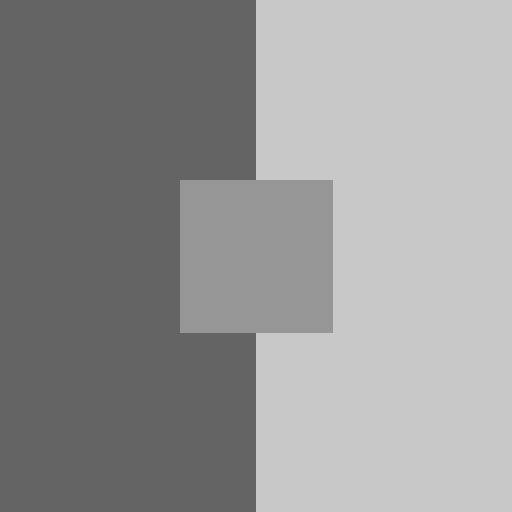}}}
\hspace{0.5cm}
\subfloat[][The MSHR of (a)]{{\fbox{\includegraphics[width=0.09\textwidth]{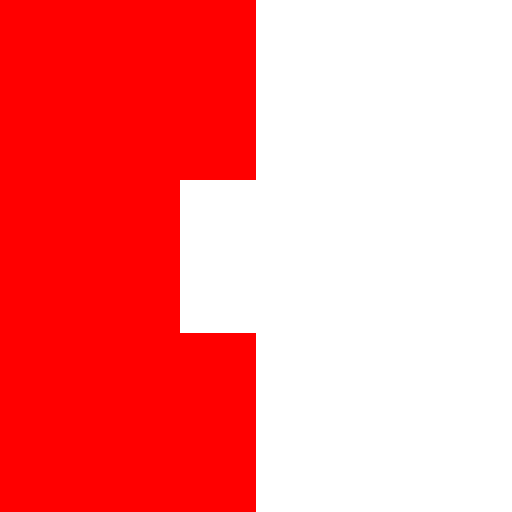}}}
{\fbox{\includegraphics[width=0.09\textwidth]{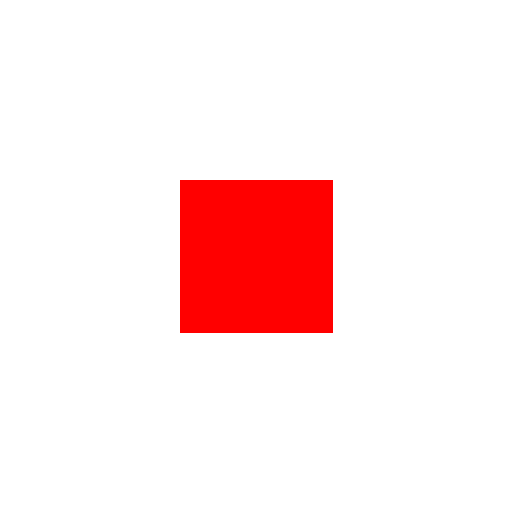}}}
{\fbox{\includegraphics[width=0.09\textwidth]{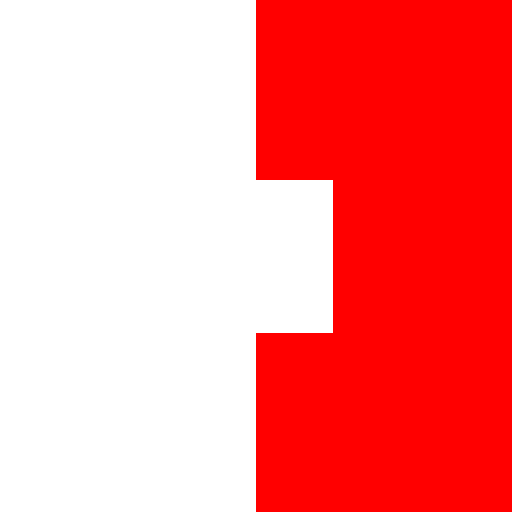}}}}
\end{center}
\begin{center}
\subfloat[][The MSER of (a)]{{\fbox{\includegraphics[width=0.09\textwidth]{NonMSERColor1.png}}}{\fbox{\includegraphics[width=0.09\textwidth]{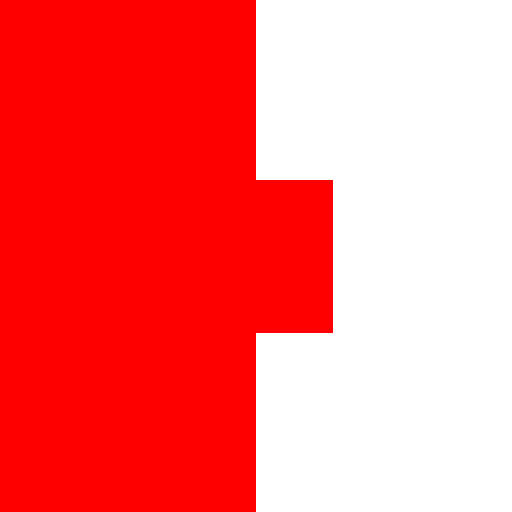}}}{\fbox{\includegraphics[width=0.09\textwidth]{NonMSERColor3.png}}}{\fbox{\includegraphics[width=0.09\textwidth]{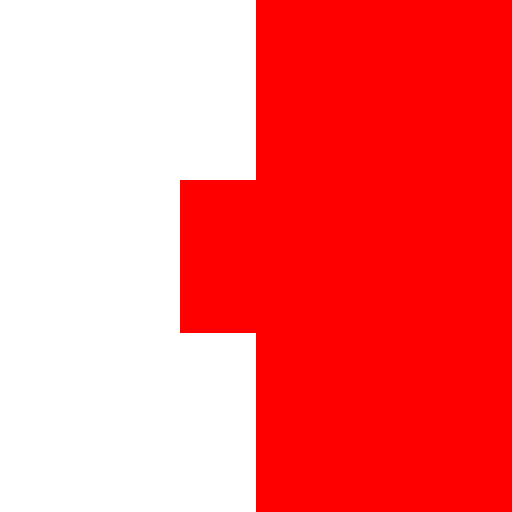}}}}
\end{center}
\caption{The center region of (a) is no extremal region since it is lighter \emph{and} darker than its background. Hence, regardless of the parameter settings, it will never be extracted as an MSER (c). On the other hand, the inner edges of the center region are smaller than its outer edges and hence it is a homogeneous region (b).}
\label{fig:nomser}
\end{figure}

To overcome this limitation, we make use of the recently proposed edge-based component-tree \cite{boettger2017gcpr}. The edge-based component-tree is an efficient extension of the component-tree devoted to gray-scale images to images with an arbitrary number of channels. The only difference is that the tree nodes do not consist of extremal regions (regions which have a gray-value strictly larger or smaller than their neighboring pixels) but of homogeneous regions. The homogeneous regions are characterized by the fact that each pixel within the region has a vertical or horizontal edge with a smaller magnitude than all outer edges of the region, please see \figref{fig:nomser} for details. By traversing the edge-based component-tree, it is possible to extract Maximally Stable Homogeneous Regions (MSHR) regions. They share the same parameters as MSER and scale equally well with growing image sizes. Using the edge-based component-tree has the further advantage that the computational complexity of the tracking stage can be significantly reduced. 

The main contribution of this paper is the efficient extension of the MSER tracking algorithm of Donoser and Bishop \cite{donoser2006efficient} to images with an arbitrary number of channels. The proposed approach opens the door to a number of new applications and scales very well to images with a large number of channels.

\section{Related Work}
For MSER tracking in gray-scale images, the component-tree is essential to enable an efficient computation and improve the robustness \cite{donoser2006efficient}. First of all, the component-tree is required to extract the MSER from the tracking target in the initialization step. Secondly, in the tracking step, the MSER are localized within the component-tree of the new frame by matching them to the nodes in the tree, where each node represents an extremal region within the new frame. This makes the approach much faster and more robust than merely matching the initial MSER to the MSER in the new frame. For more details, please see  \cite{donoser2006efficient}.

There are essentially three different kinds of component-tree computation algorithms: immersion algorithms, flooding algorithms and merge-based algorithms. Recently, Carlinet and G\'eraud \cite{carlinet2013comparison} presented an extensive comparison of the main approaches and showed that the flooding-based approaches of Wilkinson \cite{wilkinson2011fast}, Salembier \etal \cite{salembier1998antiextensive} and Nist\'er and Stew\'enius \cite{nister2008linear} are superior in terms of speed for 8-bit and 16-bit images. Hence, the natural choice to construct the component-tree in MSER tracking was a flooding-based approach \cite{donoser2006efficient}.

To the best of our knowledge, an extension of MSER tracking to multichannel images does not exist to date. This is mainly due to the fact that most extensions of MSER to color do not construct a component-tree.  For example, Chavez and Gustafson \cite{chavez2011color} transform the RGB image to the HSV color space and extract gray-value MSER on the single channels separately. Forss\'en \cite{forssen2007maximally} extract so-called Maximally Stable Color Regions (MSCR) by using pixel differences in RGB images, as opposed to the RGB values directly. Although no component-tree is constructed in the process, the idea of using pixel differences is appealing, since it does not require a user-defined partial ordering and can further be trivially extended to images with an arbitrary number of channels.

General extensions of gray-value component-trees to multi-channel component-trees have been proposed as well. To this extent, Passat and Naegel \cite{passat2014component} introduce the concept of component graphs. Unfortunately, the multi-channel component graph construction requires a suitable, user-defined, piecewise ordering of the multi-channel image data that is specific to the targeted application. In general, the approach is too specific and computationally complex to be used in a real-time tracking scenario. 

A further general extension of the component-tree to multi-channel images, which is conceptionally similar to ours, is the Multivariate tree of shapes (MToS) \cite{carlinet2015mtos}. The MToS is a five step process which first computes a tree of shapes (ToS) for each channel individually. Hence, the runtime has a large linear factor in the number of image channels. Again, computing the ToS for each channel individually is computationally demanding. We compared the edge-based tree-construction to the binaries of MToS and are approximately 10-15 times faster for a three channel image. We expect this performance advantage to be even more prominent for hyperspectral images, which contain significantly more channels.

We extend the MSER tracking to multichannel images by using the edge-based component-tree \cite{boettger2017gcpr}. In contrast to the above mentioned approaches, the edge-based component-tree is applicable to images of different domains and numbers of channels. Furthermore, it does not require any pre-defined partial ordering. The flooding-based immersion allows an efficient computation that is linear in the number of pixels and scales favorably in the number of channels. Like for the MSER tracking, the component-tree is essential to determine the to-be-tracked regions in the initialization and to efficiently compute the data association in the tracking step.

\section{MSHR Tracking}

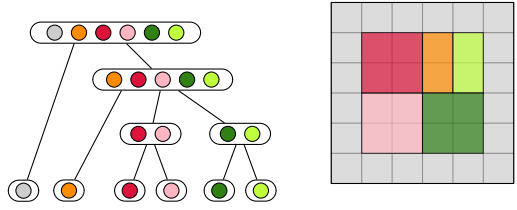
\begin{figure}
\centering
\begin{tikzpicture}[scale=0.4, every node/.style={scale=0.8}]
\begin{scope}
    \node[TreeRed] at (-3.5,0) (A) {};
    \path (A) ++(0:4mm) node [TreeBox] [minimum width = 28mm] (AA) {};
    \path (A) ++(0:16mm) node [TreeGreen] (Z) {};
    \path (A) ++(0:24mm) node [TreeLightGreen] (Z) {}; 
    \path (A) ++(180:8mm) node [TreeOrange] (Z) {};
    \path (A) ++(180:16mm) node [TreeGray] (Z) {};
    \path (A) ++(0:8mm) node [TreePink] (Z) {};
    
    \path (AA) ++(-45:22mm) node [TreePink] (B) {};
    \path (B) ++(0:0mm) node [TreeBox] [minimum width = 23mm] (BB) {};
    \draw (AA) -- (BB) node [left] {}(A);
    \path (B) ++(0:8mm) node [TreeGreen] (Z) {};
    \path (B) ++(0:16mm) node [TreeLightGreen] (Z) {}; 
    \path (B) ++(180:8mm) node [TreeRed] (Z) {};
    \path (B) ++(180:16mm) node [TreeOrange] (Z) {}; 
    \path (AA) ++(-120:60.5mm) node [TreeGray] (J) {};
    \path (AA) ++(-120:60.5mm) node [TreeBox] [minimum width = 5mm] (J) {};
    \draw (AA.195) -- (J) node [left] {}(A);    
    
    \path (B) ++(-90:18mm) node [TreePink] (C) {};
    \path (C) ++(180:8mm) node [TreeRed] (Z) {};   
    \path (Z) ++(0:4mm) node [TreeBox] [minimum width = 10mm] (CC) {};
    \draw (BB) -- (CC) node [left] {}(A);
    \path (B) ++(-40:28mm) node [TreeGreen] (D) {};
    \path (D) ++(0:4mm) node [TreeBox] [minimum width = 10mm] (DD) {};
    \path (D) ++(0:8mm) node [TreeLightGreen] (Z) {};    
    \draw (BB) -- (DD) node [left] {}(A);
    \path (BB) ++(-130:48mm) node [TreeOrange] (I) {};
    \path (BB) ++(-130:48mm) node [TreeBox] [minimum width = 5mm] (I) {};
    \draw (BB.195) -- (I) node [left] {}(A);
    
    \path (DD) ++(-110:20mm) node [TreeGreen] (E) {};
    \path (DD) ++(-110:20mm) node [TreeBox] [minimum width = 5mm] (EE) {};
    \draw (DD) -- (EE) node [left] {}(A);
    \path (DD) ++(-70:20mm) node [TreeLightGreen] (G) {};
    \path (DD) ++(-70:20mm) node [TreeBox] [minimum width = 5mm] (EE) {};
    \draw (DD) -- (EE) node [left] {}(A);
        
    \path (CC) ++(-110:20mm) node [TreeRed] (F) {};
    \path (CC) ++(-110:20mm) node [TreeBox] [minimum width = 5mm] (FF) {};
    \draw (CC) -- (FF) node [left] {}(A);
    \path (CC) ++(-70:20mm) node [TreePink] (H) {};
    \path (CC) ++(-70:20mm) node [TreeBox] [minimum width = 5mm] (FF) {};
    \draw (CC) -- (FF) node [left] {}(A);  
\end{scope}

\begin{scope}[xshift=4cm]
\draw (0, -5) grid (6,1);
	\filldraw[fill=myGray, opacity=0.7] (0, -5) rectangle (6,1);
	\filldraw[fill=myPink, opacity=0.7] (1, -4) rectangle (3,-2);
	\filldraw[fill=myRed, opacity=0.7] (3, -2) rectangle (1,0);
	\filldraw[fill=myOrange, opacity=0.7] (3, -2) rectangle (4,0);
	\filldraw[fill=myLightGreen, opacity=0.7] (4, 0) rectangle (5,-2);
	\filldraw[fill=myGreen, opacity=0.7] (3, -2) rectangle (5,-4);
\end{scope}
    
\end{tikzpicture}
\caption{Example of how the edge-based component-tree is constructed for a three-channel image (best viewed in color). In a first step, each of the uniquely colored regions is merged into a single component (the inner edge distances are 0). The most similar colors are pink and red and light and dark green. Hence, in the next step, these colors are merged. The orange region has similar distances to the red and green regions and is thus merged with the green and red components in the next step. Finally, the gray region, having the largest distance to all the colors, is merged and the complete image is connected within the edge-based component-tree.}
\label{fig:componenttree}
\end{figure}

Object tracking is typically divided into two stages, the initialization and the tracking stage. In the first step, the object location is given and the tracker is initialized. In the second step, given a new frame, the tracker identifies the most probable object location. 

To date, most trackers represent the object location as an axis-aligned box. Furthermore, most of the existing tracking benchmarks, such as VOT2016 \cite{vot_2016} or OTB \cite{wu_otb_2015}, also represent objects with axis-aligned or oriented boxes. In MSHR tracking, on the other hand, we represent the object as a dense, by-pixel, segmentation. This enables a much more precise localization of the object and allows the object to undergo non-linear deformations. Nevertheless, we require the object to be representable by a homogeneous region. 
\begin{figure*}[t]
\centering
\includegraphics[trim={0 4.8cm 0 4.4cm},clip,width=0.95\textwidth]{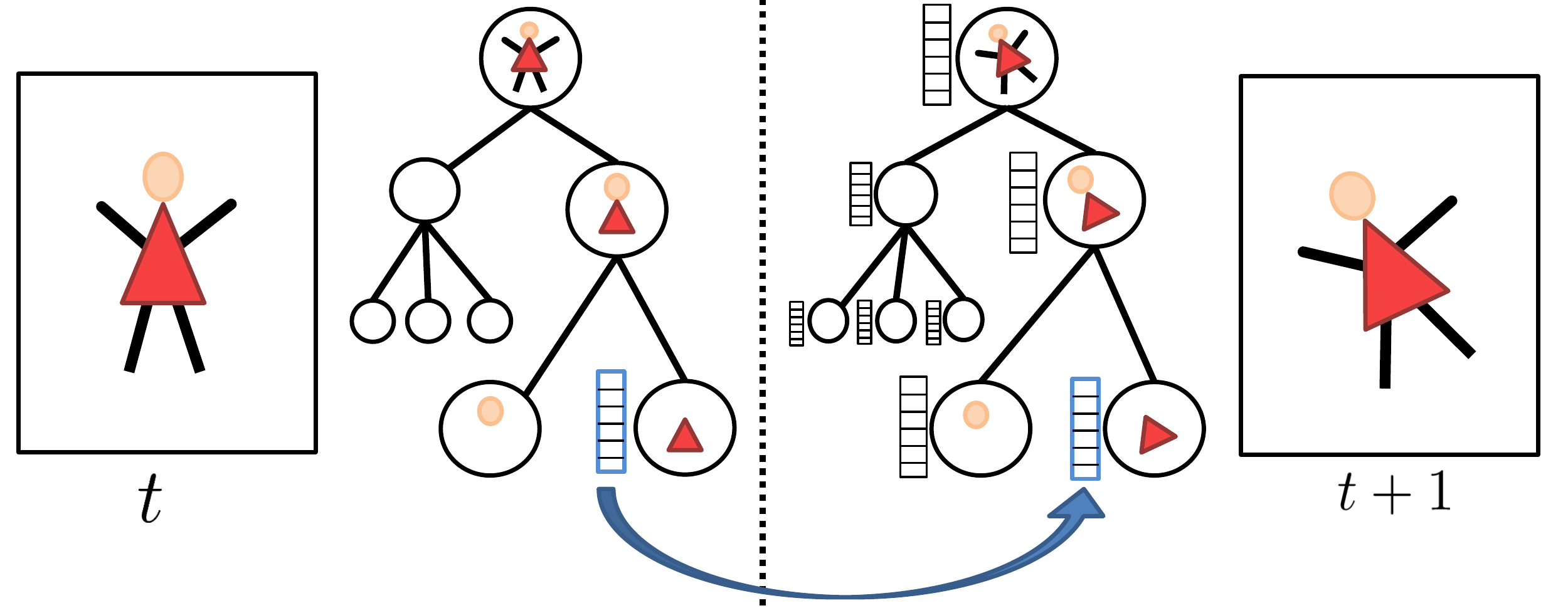}
\caption{Flowchart of the proposed MSHR tracking approach. At time step $t$, the features described in section \ref{sec:tracking} are calculated for the currently tracked MSHR. In the next time step, merely the edge-based component-tree is constructed and the features calculated for every node. These features are then matched to those of the MSHR features from time step $t$.}
\label{fig:flowchart}
\end{figure*}
\subsection{Initialization}
In a first step, we extract MSHR from the given target location. MSHR divide the image into multiple, possibly overlapping, connected components. The connectedness is not defined through the binary masks of gray-value thresholds, but rather by thresholds of edge magnitudes. The resulting components are characterized by the fact that each pixel within the region has a vertical or horizontal edge that is smaller than the current threshold and, vice versa, that all outer edges of the component are larger than the current edge magnitude threshold. By successively increasing the threshold, the components grow and eventually merge into other components. The merge processes are recorded in the component-tree. An exemplary construction of an edge-based component-tree is visualized in the toy example in \figref{fig:componenttree}. The construction process follows a flooding-based immersion and is described in more detail in \cite{boettger2017gcpr}.

Similar to MSER, a Maximally Stable Homogeneous Region $R_{i*}$ is a region that has a local \mbox{minimum} of 
\begin{equation} \label{eq:stability}
s(i)=|R_{i+\Delta }\setminus R_{i-\Delta }|-|R_{i}|,
\end{equation}
at $i*$. Here $\Delta$ is the stability parameter of the method and $|\cdot|$ denotes the cardinality. Hence, in the initialization step, the locally most stable regions are extracted from the target and selected for tracking.

\subsection{Tracking Step} \label{sec:tracking}

Given the object location in the prior frame, we first define a suitable search domain in the current frame. In all of our experiments we use a rectangular domain with twice the object bounding box extents as search region. 

To enable the matching of the MSHR in the tracking step, we compute a handful of region and gray-value features for each MSHR. We solely use features that can efficiently be calculated by region and gray-value moments \cite{steger2008machine}. The moment of order $(p,q)$ of a region $\mathcal{R}$ is defined as
\begin{equation}
m_{p,q} = \sum_{(r,c)\in \mathcal{R}} r^p c^q,
\end{equation}
where $p\geq0$ and $q\geq0$. Since the flooding-based immersion considers each image pixel in the tree construction anyway, our choice of features can be calculated while constructing the component-tree  without adding significant computational complexity. We use the area of the region ($m_{0,0}$), the center of gravity ($m_{1,1}/m_{0,0}$) and the ellipse parameters $r_1,r_2$ and $\theta$ as tracking features. The ellipse parameters can be calculated with the normalized moments, please see \cite{haralick1992lg} for details. Analogously, we use gray-value moments to calculate the average gray-value and the gray-value deviation of the single channels as further features. Please note, our selection of features makes the approach invariant to rotations of the MSHR. 

To further improve the robustness, we weight the single features in the matching step for specific applications. For example, if the object may undergo great deformations, but has a relatively constant color, the weight of the region moments can be reduced and the gray-value features weights increased. 

In the tracking step, we do not extract the most stable MSHR and match their features to those from the initial frame. Instead, we compare the features of the initial MSHR to the features all of the homogeneous regions in the component-tree nodes. This helps to improve the robustness and ensures we do not restrict the search to only \emph{maximally} stable homogeneous regions. The general idea of MSHR tracking is visualized in \figref{fig:flowchart}.
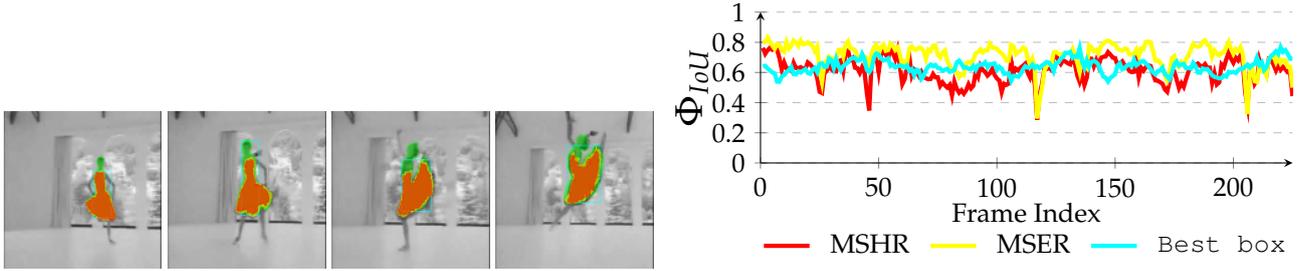
\begin{figure}[t]
{\flushright
\includegraphics[width=0.115\textwidth]{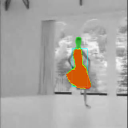}
\includegraphics[width=0.115\textwidth]{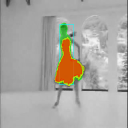}
\includegraphics[width=0.115\textwidth]{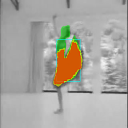}
\includegraphics[width=0.115\textwidth]{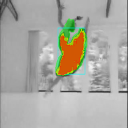}
}
\centering
\begin{tikzpicture}
\begin{axis}[
	scale only axis=true,
	width=7cm, height=2.0cm,
	x label style={at={(axis description cs:0.5,-0.2)},anchor=north},
	y label style={at={(axis description cs:-0.08,.5)},anchor=south},
    xlabel={Frame Index},
    ylabel={\Large$\Phi_{IoU}$},
    xmin=0, xmax=225,
    ymin=0, ymax=1.00,
    xtick={0,50,100,150,200},
    ytick={0,.20,.40,.60,.80,1.00},
    legend style={at={(0.5, -0.4)}, anchor = north, column sep=5pt, draw=none},
    legend columns=3,
    ymajorgrids=true,
    grid style=dashed,
    axis lines=left,
    every axis plot/.append style={ultra thick}
]
 
\addplot [color=colorMSHR] table {dress_overlap.dat};
\addplot [color=colorMSER] table {dress_overlap_mser.dat};
\addplot [color=colorBox] table {dress_overlap_box.dat};
\legend{MSHR, MSER, {\tt{Best box}}}
\end{axis}
\end{tikzpicture}
\hfill
\caption{{\tt dress} from OTB \cite{wu_otb_2015}. For this gray scale scene, the MSER tracker is able to outperform the MSHR tracker. The tracker also clearly outperforms the best possible overlap an axis-aligned tracker ({\tt{Best box}}) can achieve for the segmentations within the scene.}
\label{fig:dress_example}
\end{figure}

\subsection{Model Update}

To enable robust tracking, as opposed to \cite{donoser2006efficient}, we update the region features incrementally in each frame. This enables to handle short occlusions and detection failures in single frames. Hence, after successfully locating the node that best fits the to-be-tracked MSHR, we update the feature vector as

\begin{equation}
\label{eq:update}
\text{feat}_{t+1} = (1-\lambda) \text{feat}_t + \lambda \text{feat}_{t+1}.
\end{equation}
In our experiments we used $\lambda = 0.5$.
\section{Evaluation}

The proposed MSHR tracking approach is not restricted to bounding boxes. Hence, to evaluate the quality of the tracking results, we manually annotated dense by-pixel segmentations of a handful of scenes from the OTB \cite{wu_otb_2015} and VOT2016 \cite{vot_2016} datasets. Otherwise, the given bounding box groundtruth would introduce an undesired bias when measuring the overlap scores of by-pixel segmentations. As accuracy measure, we use the commonly used Intersection over Union (IoU) criterion. The IoU of a tracker proposition $\mathcal{B}$ and the groundtruth segmentation $\mathcal{S}$ is computed as,
\begin{equation}
\label{eq:IoU}
\Phi_{IoU} \left( \mathcal{S}, \mathcal{B}\right ) = \frac{\left \vert \mathcal{S} \cap \mathcal{B} \right \vert}{\left \vert \mathcal{S} \cup \mathcal{B} \right \vert}.
\end{equation}

To bring the results into perspective, we compute the best possible overlap an axis-aligned tracker could obtain for the segmentation of a given scene. By this means, the performance gain of using segmentations can be highlighted without introducing a bias by choosing a specific set of state-of-the-art axis-aligned trackers to compete against. We refer to this tracker as the {\tt{Best box}}. The tracker is computed by optimizing
\begin{equation}
\label{eq:optimization}
\Phi_{opt}(\mathcal{S}) = \max_b\,\,\, \Phi_{IoU} (\mathcal{S},\mathcal{B}(b)) \hspace{1cm} s.t.\,\, b \in \mathbb{R}_{>0}^4,
\end{equation}
where $b$ is the axis-aligned box parametrization, which has 4 degrees of freedom. We optimize \eqref{eq:optimization} by exhaustively searching for the best axis-aligned bounding box of each segmentation.

To help understand the difference between MSER and MSHR tracking, we further compare our approach to a version of the MSER tracker \cite{donoser2006efficient}. To focus the evaluation on the different regions both approaches use, and not on their features, we use the exact same parameters and moment-based features for both approaches. 

For the gray-scale sequence {\tt dress} from OTB \cite{wu_otb_2015}, the MSER tracker  outperforms the {\tt{Best box}} and the MSHR tracker, as is displayed in \figref{fig:dress_example}. In the respective sequence, the MSER tracker is able to track the head and the dress of the dancer, while the MSHR tracker only tracks the dress. Hence, the overlap scores of MSER are superior. Nevertheless, it is important to note that both approaches are compared against the \emph{best possible} axis-aligned tracker and, accordingly, the overlap scores are impressive.

\begin{figure}
{\flushright
\includegraphics[width=0.115\textwidth]{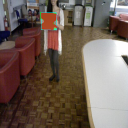}
\includegraphics[width=0.115\textwidth]{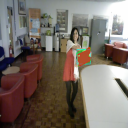}
\includegraphics[width=0.115\textwidth]{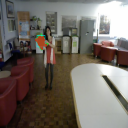}
\includegraphics[width=0.115\textwidth]{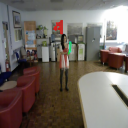}
}
\centering
\begin{tikzpicture}
\begin{axis}[
	scale only axis=true,
	width=7cm, height=2.0cm,
	x label style={at={(axis description cs:0.5,-0.2)},anchor=north},
	y label style={at={(axis description cs:-0.08,.5)},anchor=south},
    xlabel={Frame Index},
    ylabel={\Large$\Phi_{IoU}$},
    xmin=0, xmax=55,
    ymin=0, ymax=1.00,
    xtick={0,10,20,30,40,50},
    ytick={0,.20,.40,.60,.80,1.00},
    legend style={at={(0.5, -0.4)}, anchor = north, column sep=5pt, draw=none},
    legend columns=3,
    ymajorgrids=true,
    grid style=dashed,
    axis lines=left,
    every axis plot/.append style={ultra thick}
]
 
\addplot [color=colorMSHR] table {book_overlap.dat};
\addplot [color=colorMSER] table {book_overlap_mser.dat};
\addplot [color=colorBox] table {book_overlap_box.dat};

\legend{MSHR, MSER, {\tt{Best box}}}
\end{axis}
\end{tikzpicture}
\hfill
\caption{{\tt book} from VOT2016 \cite{vot_2016}. Since the gray-scale region is not an MSER, it cannot be tracked with MSER tracking (see \figref{fig:book_mshr} for details). The overlap scores of the MSHR tracking are comparable and sometimes even better than the overlap the best possible axis-aligned tracker could theoretically achieve. }
\label{fig:book_example}
\end{figure}

\begin{figure}[t]
\centering

\hspace*{0.1cm}
\includegraphics[width=0.23\textwidth]{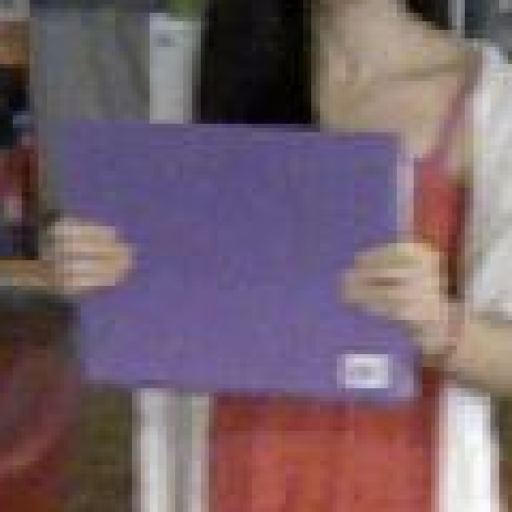}
\includegraphics[width=0.23\textwidth]{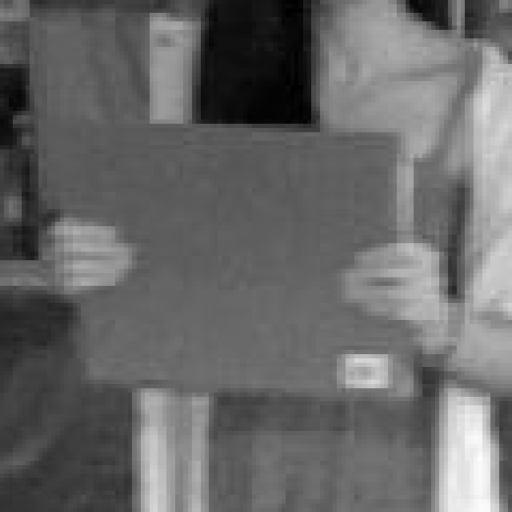}

\hfill
\caption{A close-up of {\tt book} from VOT2016 \cite{vot_2016}. Tthe book is not an MSER in the gray-scale image, since its background is lighter \emph{and} darker than the book itself. Hence, it cannot be tracked with the existing MSER tracking. The book is a homogeneous region though, and can efficiently be tracked with the proposed MSHR tracking. See \figref{fig:book_example} for the overlap scores. }
\label{fig:book_mshr}
\end{figure}

For color images, the difference of MSER and MSHR becomes more evident. In the {\tt book} sequence from VOT2016 \cite{vot_2016}, the MSER tracker fails, as is shown in \figref{fig:book_example}. The book is, per definition, not an extremal region in the  gray-scale image, as can be seen in more detail in \figref{fig:book_mshr}. Hence, the initialization is not successful and the MSER tracker fails. Nevertheless, the book is a homogeneous region in both the gray-scale and the color image, and accordingly, the MSHR tracker is successful. In most frames, the MSHR tracker is even able to outperform the {\tt{Best box}} and obtains an average IoU of $0.7$.

For the {\tt book} sequence, the MSHR tracking requires an average of 21ms  per frame and for the  
{\tt dress} sequence an average of 16ms per frame. The algorithm is implemented in HALCON\footnote{MVTec Software GmbH, \url{https://www.mvtec.com/}} and run on an IntelCore i7-4810 CPU @2.8GHz with 16GB of RAM with Windows 7 (x64).

\section{Applications}
For MSER tracking, Donoser and Bischof \cite{donoser2006efficient} presented three different applications; license plate tracking, face tracking and the segmentation of a fiber network. In the third application, a fiber network is reconstructed in 3D by tracking a slice of the data along the axis orthogonal to the image data. Analogously, we track organs in slices of a Computed Tomography (CT) scan, to generate a 3D segmentation. We use the CT data provided in the 3DIRCADb dataset\footnote{The dataset is available on
\url{http://ircad.fr/research/3d-ircadb-01}} \cite{soler_2012_database}.

\begin{figure}[t]
\centering
\subfloat[][]{\includegraphics[width=0.155\textwidth]{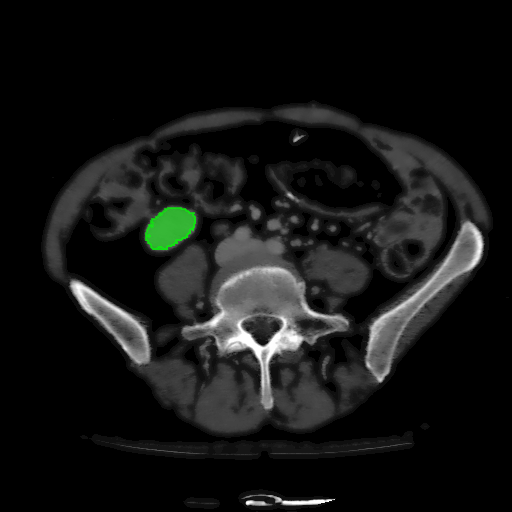}
\includegraphics[width=0.156\textwidth]{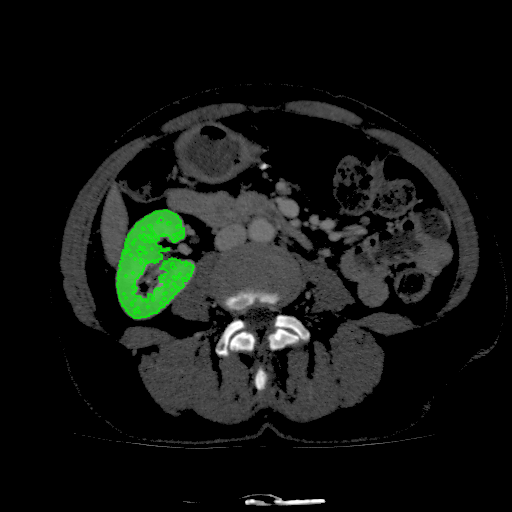}
\includegraphics[width=0.156\textwidth]{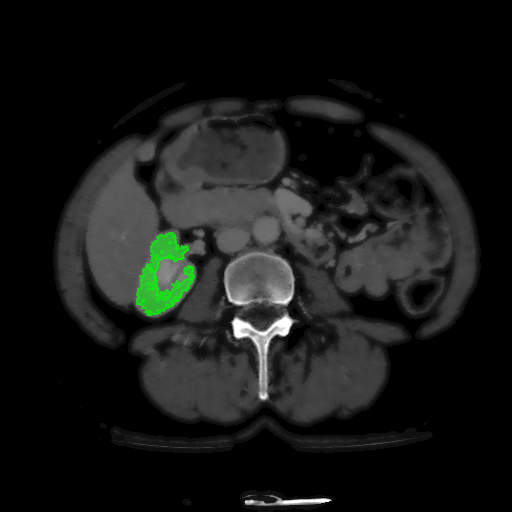}}\\
\subfloat[][]{\includegraphics[width=0.155\textwidth]{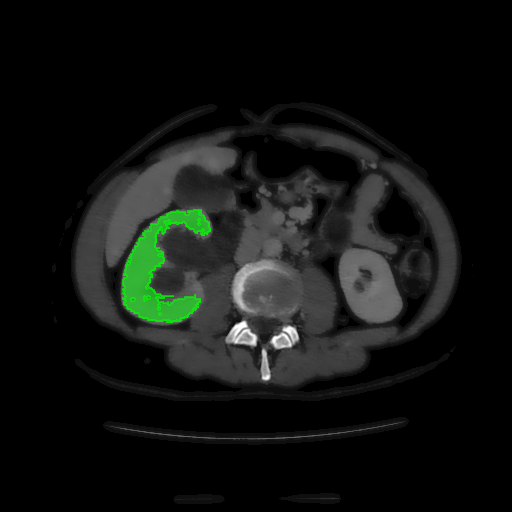}
\includegraphics[width=0.156\textwidth]{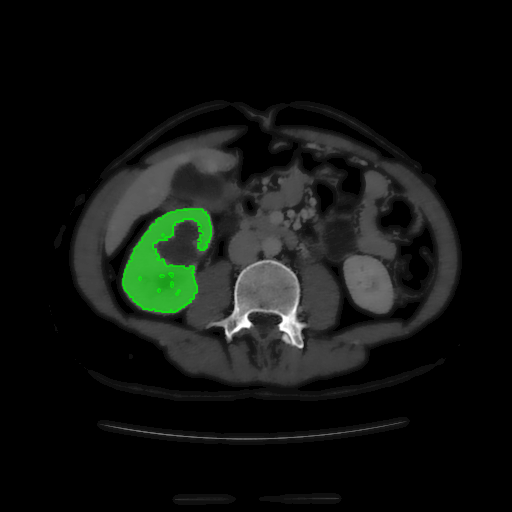}
\includegraphics[width=0.156\textwidth]{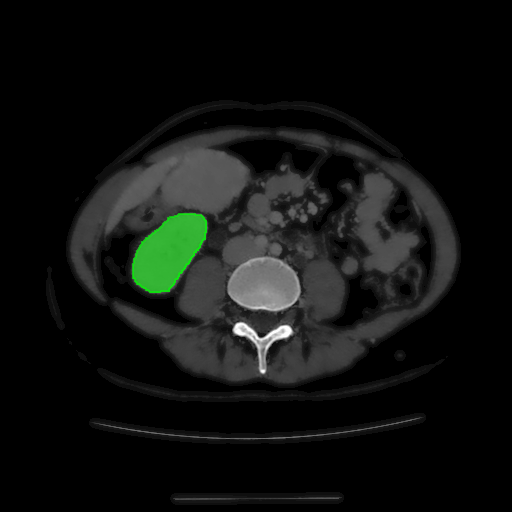}}
\caption{Two examples sequences from the 3DIRCADb dataset  \cite{soler_2012_database}. Given an initial selection of a single slice (the middle image in (a) and (b)) of the organ, the proposed MSHR tracking tracks the region forward and backwards in space. The segmented slices can be used to reconstruct the organ, see \figref{fig:reconstruction} for an example reconstruction.}
\label{fig:application}
\end{figure}

\begin{figure}[t]
\centering
\includegraphics[width=0.24\textwidth]{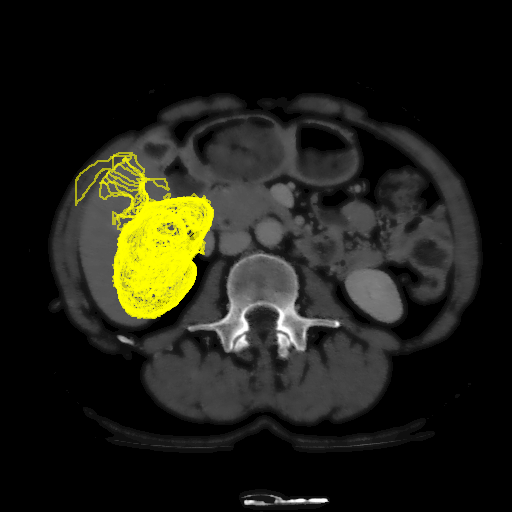}
\includegraphics[width=0.24\textwidth]{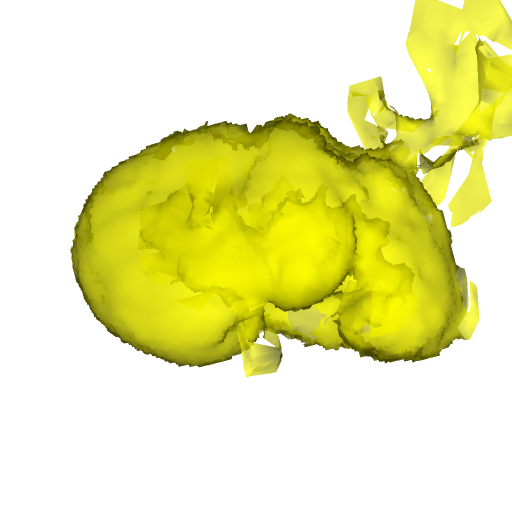}\\
\includegraphics[width=0.24\textwidth]{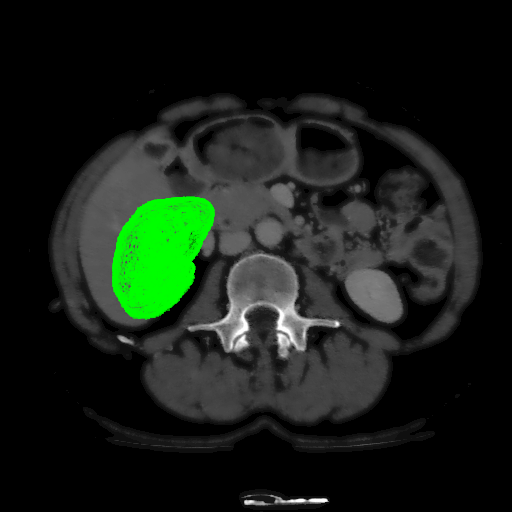}
\includegraphics[width=0.24\textwidth]{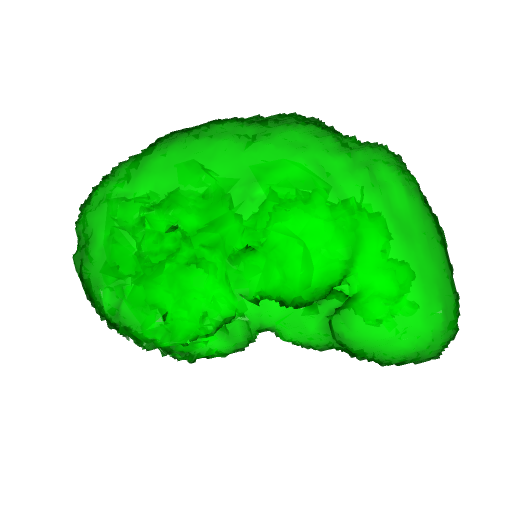}
\caption{In the first row, the reconstruction of the liver is displayed for MSER tracking. The low contrast and the fact that the background is partly darker and lighter than the objects makes the reconstruction noisy. The proposed MSHR tracking can cope with these situations and the reconstruction is significantly better.}
\label{fig:reconstruction}
\end{figure}

To initialize the tracking process, the organ is segmented in an arbitrary slice of the CT data by a bounding box. The most stable MSHR is then selected in the initialization process for tracking. The respective MSHR is tracked through the slice data along the axis orthogonal to the image data. An example of the tracked regions is visualized for two examples in \figref{fig:application}. Given the segmentations of the single slices, the organ can be reconstructed in 3D. We compare the reconstruction for MSER and MSHR tracking in \figref{fig:reconstruction}. To enhance the visualization, the datapoints are triangulated and the surface normals calculated. Since the contrast of the organs can be very low in CT images, the MSER tracking has difficulties catching the organ boundaries. Furthermore, the organ is sometimes partely lighter \emph{and} darker than the background, which may lead to MSER tracking failure. The proposed MSHR tracking copes well with these difficulties, and the reconstructions are significantly better.

Please note, the tracking of the regions in the CT slices is extremely efficient and only requires an average of 5ms per slice. Hence, for the 45 slices in \figref{fig:reconstruction} the complete 3D reconstruction process, which includes the triangulation ($\approx 1s$), the calculation of the surface normals ($\approx130ms$), and the segmentation ($\approx220ms$), requires only around 1.5s.

\section{Conclusion}
We presented an efficient extension of the MSER tracking algorithm to images with an arbitrary number of channels. We evaluated the approach on a collection of densely segmented ground-truth sequences and display its advantages to MSER tracking. The approach is linear in the number of pixels and runs at 50 fps on the evaluated tracking sequences. Furthermore, the approach was used for the 3D reconstruction of organs in CT images. It was more accurate compared to the original MSER tracking and has a very low computational complexity.

\bibliography{mshrbib}

\end{document}

%% file: preamble.tex
\usepackage[a4paper,textwidth=18cm,textheight=24cm,top=2.85cm, bottom=2.85cm, left=1.5cm, right=1.5cm]{geometry}

\usepackage{icdp2009}

\makeatletter
\long\def\@makecaption#1#2{%
\vskip\abovecaptionskip
\sbox\@tempboxa{#1. #2}%
\ifdim \wd\@tempboxa >\hsize
#1. #2\par
\else
\global \@minipagefalse
\hb@xt@\hsize{\box\@tempboxa\hfil}%
\fi
\vskip\belowcaptionskip}
\makeatother

%% file: main.bbl
\begin{thebibliography}{10}

\bibitem{boettger2017gcpr}
T.~B{\"{o}}ttger and D.~Gutermuth.
\newblock Edge-based component-trees for multi-channel image segmentation.
\newblock {\em CoRR}, abs/1705.01906, 2017.

\bibitem{carlinet2013comparison}
E.~Carlinet and T.~G{\'{e}}raud.
\newblock A comparison of many max-tree computation algorithms.
\newblock In {\em Mathematical Morphology and Its Applications to Signal and
  Image Processing, 11th International Symposium}, pages 73--85, 2013.

\bibitem{carlinet2015mtos}
E.~Carlinet and T.~G{\'e}raud.
\newblock Mtos: A tree of shapes for multivariate images.
\newblock {\em IEEE Transactions on Image Processing}, 24(12):5330--5342, 2015.

\bibitem{chavez2011color}
A.~Chavez and D.~Gustafson.
\newblock Color-based extensions to {MSERs}.
\newblock In {\em Advances in Visual Computing - 7th International Symposium,
  {ISVC}}, pages 358--366, 2011.

\bibitem{donoser2006efficient}
M.~Donoser and H.~Bischof.
\newblock Efficient maximally stable extremal region {(MSER)} tracking.
\newblock In {\em {IEEE} Conference on Computer Vision and Pattern
  Recognition}, pages 553--560, 2006.

\bibitem{forssen2007maximally}
P.~Forss{\'{e}}n.
\newblock Maximally stable colour regions for recognition and matching.
\newblock In {\em {IEEE}Conference on Computer Vision and Pattern Recognition},
  pages 1--8, 2007.

\bibitem{haralick1992lg}
R.~M. Haralick and L.~G. Shapiro.
\newblock {\em Computer and Robot Vision, Vol. 1}.
\newblock Addison-Wesley Longman Publishing Co., Inc., 1991.

\bibitem{vot_2016}
M.~Kristan, A.~Leonardis, J.~Matas, M.~Felsberg, R.~P. Pflugfelder, L.~Cehovin,
  T.~Voj{\'{\i}}r, and G.~H{\"{a}}ger.
\newblock The visual object tracking {VOT2016} challenge results.
\newblock In {\em {ECCV} Workshops}, pages 777--823, 2016.

\bibitem{matas2004robust}
J.~Matas, O.~Chum, M.~Urban, and T.~Pajdla.
\newblock Robust wide baseline stereo from maximally stable extremal regions.
\newblock In {\em Proceedings of the British Machine Vision Conference 2002},
  pages 1--10, 2002.

\bibitem{neumann2012real}
L.~Neumann and J.~Matas.
\newblock Real-time scene text localization and recognition.
\newblock In {\em {IEEE} Conference on Computer Vision and Pattern
  Recognition}, pages 3538--3545, 2012.

\bibitem{nister2008linear}
D.~Nist{\'{e}}r and H.~Stew{\'{e}}nius.
\newblock Linear time maximally stable extremal regions.
\newblock In {\em Computer Vision - European Conference on Computer Vision},
  pages 183--196, 2008.

\bibitem{passat2014component}
N.~Passat and B.~Naegel.
\newblock Component-trees and multivalued images: Structural properties.
\newblock {\em Journal of Mathematical Imaging and Vision}, 49(1):37--50, 2014.

\bibitem{salembier1998antiextensive}
P.~Salembier, A.~Oliveras{-}Verg{\'{e}}s, and L.~Garrido.
\newblock Antiextensive connected operators for image and sequence processing.
\newblock {\em {IEEE} Transactions on Image Processing}, 7(4):555--570, 1998.

\bibitem{soler_2012_database}
L.~Soler, A.~Hostettler, V.~Agnus, A.~Charnoz, J.B. Fasquel, J.~Moreau,
  A.~Osswald, M.~Bouhadjar, and J.~Marescaux.
\newblock 3d image reconstruction for comparison of algorithm database: a
  patient-specific anatomical and medical image database, 2010.

\bibitem{steger2008machine}
C.~Steger, M.~Ulrich, and C.~Wiedemann.
\newblock {\em Machine vision algorithms and applications}.
\newblock Wiley-VCH, Weinheim, 2007.

\bibitem{wilkinson2011fast}
M.~H.~F. Wilkinson.
\newblock A fast component-tree algorithm for high dynamic-range images and
  second generation connectivity.
\newblock In {\em 18th {IEEE} International Conference on Image Processing},
  pages 1021--1024, 2011.

\bibitem{wu_otb_2015}
Y.~Wu, J.~Lim, and M.~Yang.
\newblock Object tracking benchmark.
\newblock {\em {IEEE} Transactions on Pattern Analysis and Machine
  Intelligence}, 37(9):1834--1848, 2015.

\end{thebibliography}
